\documentclass[conference]{IEEEtran}
\IEEEoverridecommandlockouts
\usepackage{cite}
\usepackage{amsmath,amssymb,amsfonts}
\usepackage{algorithmic}
\usepackage{graphicx}
\usepackage{textcomp}
\usepackage{xcolor}
\usepackage{multirow}
\usepackage{booktabs}
\usepackage{blindtext}
\usepackage{hyperref}

\def\BibTeX{{\rm B\kern-.05em{\sc i\kern-.025em b}\kern-.08em
    T\kern-.1667em\lower.7ex\hbox{E}\kern-.125emX}}
\begin{document}

\title{XFL: A High Performace, Lightweighted Federated Learning Framework}

\author{\IEEEauthorblockN{1\textsuperscript{st} Hong Wang}
\IEEEauthorblockA{\textit{\qquad\qquad Basebit.ai \qquad\qquad} \\
hong.wang@basebit.ai}
\and
\IEEEauthorblockN{2\textsuperscript{nd} Yuanzhi Zhou}
\IEEEauthorblockA{\textit{\qquad\qquad Basebit.ai \qquad\qquad} \\
yuanzhi.zhou@basebit.ai}
\and
\IEEEauthorblockN{3\textsuperscript{rd} Chi Zhang}
\IEEEauthorblockA{\textit{\qquad\qquad Basebit.ai \qquad\qquad} \\
chi.zhang@basebit.ai}
\and
\IEEEauthorblockN{4\textsuperscript{th} Chen Peng}
\IEEEauthorblockA{\textit{\qquad\qquad Basebit.ai \qquad\qquad} \\
chen.peng@basebit.ai}
\and
\IEEEauthorblockN{5\textsuperscript{th} Mingxia Huang}
\IEEEauthorblockA{\textit{\qquad\qquad Basebit.ai \qquad\qquad} \\
mingxia.huang@basebit.ai}
\and
\IEEEauthorblockN{6\textsuperscript{th} Yi Liu}
\IEEEauthorblockA{\textit{\qquad\qquad Basebit.ai \qquad\qquad} \\
yi.liu@basebit.ai}
\and
\IEEEauthorblockN{7\textsuperscript{th} Lintao Zhang}
\IEEEauthorblockA{\textit{\qquad\qquad Basebit.ai \qquad\qquad} \\
lintao.zhang@basebit.ai}
}

\maketitle

\begin{abstract}
 This paper introduces XFL, an industrial-grade federated learning project. XFL supports training AI models collaboratively on
 multiple devices, while utilizes homomorphic encryption, differential privacy, secure multi-party computation and other 
 security technologies ensuring no leakage of data. XFL provides an abundant algorithms library, integrating a large number 
 of pre-built, secure and outstanding federated learning algorithms, covering both the horizontally and vertically federated 
 learning scenarios. Numerical experiments have shown the prominent performace of these algorithms. XFL builds a concise 
 configuration interfaces with presettings for all federation algorithms, and supports the rapid deployment via docker containers.
 Therefore, we believe XFL is the most user-friendly and easy-to-develop federated learning framework. XFL is open-sourced, and 
 both the code and documents are available at https://github.com/paritybit-ai/XFL.
\end{abstract}
  
  
\section{Introductio}
 Data privacy and security has been widely awared of an important field to the wide application of artifical intelligence(AI).
 Completing model training and inference under the premise of good privacy and data management is the key to expand the value of 
 AI technology. Since McMahan et al\cite{FedAvg} proposed federated learning, it has been one of the most well-known solutions for 
 secure and privacy-preserving computing\cite{Kairous2019, QYang2019}. Federated learning attempts to utilize data from multiple 
 devices or silos to train AI models jointly. To make use of data distributed among multiple participants, traditional machine 
 learning algorithms need to be adapted to the federated scenes. Besides, to meet the requirements of privacy computing, machine 
 learning need to be reasonably combined with encryption methods. Therefore, a mature framework must have an algorithm library 
 covering comprehensive machine learning and encryption algorithms, which are exactly scarce in existing frameworks. Furthermore, 
 most federated learning frameworks are difficult for industrial production because they involves obscure knowledge and too many 
 hardware and software dependencies. The above problems have become a huge gap between the federated learning framework and a data 
 scientist who wants to apply it to a production.
 
 According to the data partition methods, there are usually two types of federated learning scenarios. The horizontal federated 
 learning scenario is that the datasets share the same feature space but differ in samples, while the vertical federated learning 
 refers to the scenario where the datasets are splitted by features. Many frameworks like TFF\cite{TFF}, PySyft\cite{PySyft} and 
 Flower\cite{Flower} have been proposed to deal with horizontal federated learning inspired by FedAvg\cite{FedAvg}. There are also 
 some great efforts to build an all-in-one framework covering both horizontal and vertical scenarios, such as FATE\cite{FATE} and 
 FedML\cite{FedML}. However, some benchmarks\cite{LEAF, UniFed} have shown the performace of existing federated learning 
 algorithms and frameworks. Differences in performance and functionality prevent these frameworks from being widely used. In 
 addition, since most federated learning frameworks are developed based on academic research, powerful industrial-grade frameworks 
 are rare. For example, in the horizontal scenario, the aggregation algorithms implemented in the most of existing frameworks is 
 FedAvg, the original and simplest aggregation algorithm. Moreover, some frameworks can only accommodate two participants for 
 calculations, and cannot operate on large-scale datasets.

 \begin{figure*}[t]
  \centering
  \includegraphics*[width=160mm]{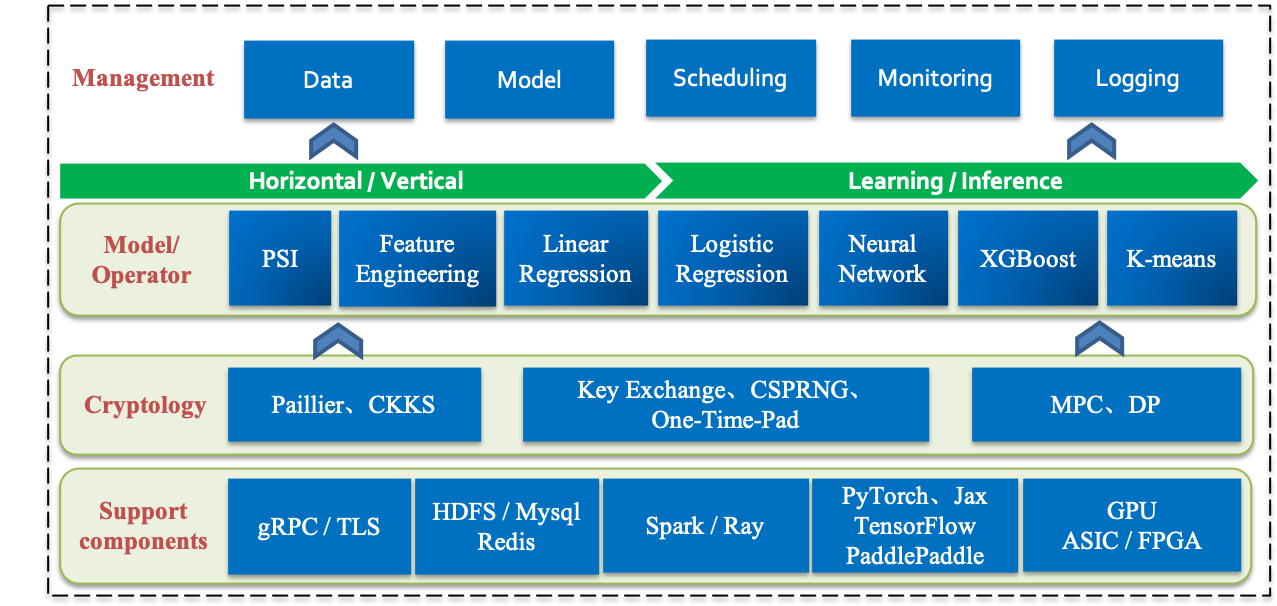}
  \caption{Architecture of XFL.}
  \label{architecture}
 \end{figure*}
  
 To build a bridge between AI and data, we introduce XFL: a high-performance, high-flexibility, high-applicability, lightweight, 
 open and easy-to-use federated learning framework. To meet the needs of privacy protection, one federated learning platform 
 must avoid direct and indirect leakage of users' local data, and enable data scientists to train model legally and compliantly. 
 XFL applies secure communication protocols to ensure communication security, and supports a range of encryption algorithms: 
 homomorphic encryption, differential privacy, secure multi-party computation and other security technologies. Unavoidably, in 
 federated learning tasks, data scientists need to make trade-offs between security and efficiency, and the training process 
 inevitably involves communication bottleneck. XFL is optimized in these aspects and has outstanding performance in every 
 federation scenario. XFL can also tolerate high latency and frequent packet loss to cope with various unstable network 
 environments. XFL inclines to be the most user-friendly federated learning framework to help popularize privacy-preserving 
 techniques in machine learning. XFL is designed in an open and extensible way, which supports deployment on CPU and GPU, and can 
 be quickly deployed via docker. Deep learning frameworks Pytorch\cite{Pytorch}, Tensorflow\cite{Tensorflow}, 
 PaddlePaddle\cite{PaddlePaddle} and Jax\cite{Jax} are supported, and the XFL library contains all mainstream horizontal/vertical 
 federation algorithms to help data scientists solve training tasks in specific scenarios. Besides, data scientists can easily 
 develop their own algorithms or models to deal with highly customized problems, due to the lightweight and superior framework 
 design of XFL.

 This paper consists of the following parts: Section \ref{Overview} introduces the architecture of XFL and the possible roles of 
 participants in a training task. Section \ref{Features} lists the main advantages and features of XFL, and in section 
 \ref{Performance}, we designed some experiments to show the outstanding performance of XFL. Section \ref{Conclusion} is the 
 conclusion of this paper.

\section{Overview of XFL}\label{Overview}

\subsection{Architecture}
 From a global perspective, XFL mainly consists of 4 layers. As the architecture shown in Fig. \ref{architecture}, from bottom to 
 top the 4 layers are:

\subsubsection{Support components}
 This layer is the cornerstone on which we build the XFL framework. The widely used PRC framework, gRPC is used to support secure 
 communication, and redis, which is an in-memory high-throughput database, is used to construct a precise API for federated 
 learning communication process. HDFS and MySql are supported for dataset storage. Spark is used for large scale data computing 
 for some certain algorithms. XFL is compatible with the mainstream deep learning frameworks such as Pytorch, Tensorflow, 
 PaddlePaddle and Jax, meanwhile it supports hardwares such as GPU, ASIC or FPGA to accelerate the computation in federated 
 learning.

\subsubsection{Cryptograhic components}
 Definitely, cryptography is one of the core modules in federated learning. XFL integrates several homomorphic encryption 
 algorithms, including Paillier\cite{Paillier} developed by us, and CKKS\cite{CKKS,TenSeal,Seal}. Secure aggregation methods for 
 horizontal federated learning contain secure key exchange, secure presudo-random number generator and one-time-pad. Differential 
 privacy is implemented for some algorithms. Support for MPC is in plan.

\subsubsection{Models / Operators}
 XFL presets abundant fedration algorithms, covering mainstream algorithms of supervised learning and unsupervised learning, 
 horizontal federation and vertical federation. Federated transfer learning is also in plan. A general horizontal federation 
 development framework and many secure aggregation methods such as FedAvg\cite{FedAvg}, FedProx\cite{FedProx}, 
 Scaffold\cite{Scaffold} are provided. XFL supports a variety of deep learning frameworks. Data scientists can build their own 
 horizontal federated learning model at a very small cost without concerning about the realization of communication and encryption 
 methods.

\subsubsection{Management}
 Support management of dataset and federated model. Provide a task schedulering, monitoring, and a module for logging.

\subsection{Roles}
\begin{figure}[h]
  \centering
  \includegraphics*[width=86mm]{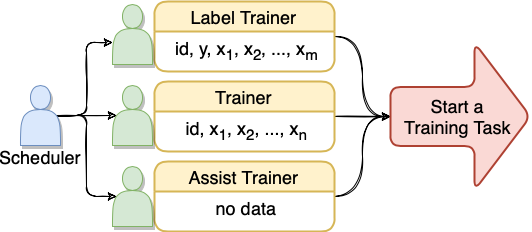}
  \caption{Roles in XFL.}
  \label{roles}
 \end{figure}

 In the federated learning process, some of the existing federated learning frameworks limit the initiator to the party who owns 
 the label in vertical scenarios. We notice in a practical application scenario, any participant should be able to initiate 
 a federated learning task. A initiator could be one of the data providers who take part in the federated learning algorithm 
 directly, or even a party who doesn't provide any data, but only intends to make use of the data from the other parties. 
 Therefore, XFL designs an independent initiator for FL tasks, which role is called scheduler as shown in Fig. \ref{roles}. The 
 scheduler is responsible for delivering the algorithm configuration and notifying the participants to start a federated learning 
 task. Apart from the scheduler, XFL defines three types of trainers. Respectively, they are label trainer, trainer and assist 
 trainer. Both the label trainer and trainer can provide the features required for model training, but the label trainer must 
 provide the labels. Supplementarily, assist trainer only do auxiliary calculations during the training process without providing 
 any data. Not all roles of trainers are needed in a federated algorithm. For example, a typical horizontal federation is consist 
 of many label trainers and one assist trainer; in the vertical scenarios, there are many trainers and one label trainer, 
 meanwhile there may also be an assist trainer or not, which depends on the concrete design of the algorithm.

\section{Features of XFL}\label{Features}
 
\subsection{Easy to Use}
 In order to encourage users to experience the XFL framework and facilitate debugging algorithms, the standalone mode is supported 
 in XFL. The standalone mode allows users to execute and test all the federated learning algorithms integrated on XFL locally, 
 just by changing only a few network configurations. Moreover, XFL provides plenty of demos corresponding to the algorithms to 
 help a user get started better and faster.

 XFL supports task orchestration, where operators can be organized in a list to execute one by one. The local operators (as shown 
 in TABLE \ref{algorithms}) can be connected to the federated operators to form a complete federated training process. For better 
 flexibility, the role of a party played in a task flow can vary in different stages, instead of being simply fixed. For instance, 
 as metioned in previous section, the role named scheduler can be changed to every one of the existing party involved in the 
 federated training task flow, or even another independent party.

 On the other hand, XFL is designed to be quite lightweight. XFL runs well in a local self-built environment, as well as in a 
 container for easy deployment and migration.

\subsection{Easy to Develop}
 To develop a self-defined algorithm, one need to define a "class" and implement a "fit" method for each role involved in the 
 federation. XFL provides vivid templates and tools to help developers build their models or operators fast and smoothly, while 
 pay as little attention to the details of communication and encryption method as possible.

 In horizontal federated learning, secure aggregation\cite{SecureAggregation} is a classic way to aggregate the gradients or 
 parameters, which is executed by an assist trainer in XFL. The templates for horizontal scenarios integrate the secure 
 aggregation flow internally, so developers don't need to care about the implementation methods. These templates could help 
 accelerate the development, by referring to which developers can define the structure of their nerual networks and the training 
 process more easily. The templates also allow developers to insert custom codes or functions by the "hook" method to enrich the 
 operator's function. To help developers with different habits run a new operator faster, XFL offers templates for various deep 
 learning framework, including Pytorch, Tensorflow, PaddlePaddle, and Jax. In addition, XFL supports several configurable aggreation 
 methods such as FedAvg, FedProx and Scaffold.

 In vertical federated learning, it is difficult to summarize a common process for most of the algorithms. Thus, developers need 
 to write encryption and commucation by their own. It is a good idea to refer to existing vertical operators in XFL before 
 developing a new operator. XFL offers cryptographic library such as paillier, key exchange protocol and some third-party tools 
 for developers to build a encryption block. A precise communcation module is also provided for developers to build a peer-to-peer 
 or broadcast channel.

 \begin{table}[b!]
  \begin{center}
    \caption{Algorithm list}
    \label{algorithms}
    \begin{tabular}{cc}
      \toprule
      Type & Algorithms \\
      \midrule
      \multirow{6}*{Horizontal} & Linear Regression \\
      ~ & Logistic Regression \\
      ~ & ResNet \\
      ~ & VGG \\
      ~ & BERT \\
      ~ & NbAFL \\
      \midrule
      \multirow{7}*{Vertical} & Binning Woe IV \\
      ~ & Pearson \\
      ~ & Feature Selection \\
      ~ & Logistic Regression\cite{SYang2019} \\
      ~ & XGBoost\cite{Secureboost} \\
      ~ & K-means \\
      ~ & Sampler \\
      \midrule
      \multirow{4}*{Local} & Normalization \\
      ~ & Standard Scaler \\
      ~ & Data Split \\
      ~ & Feature Preprocess \\
      \bottomrule
    \end{tabular}
  \end{center}
 \end{table}

 \begin{table*}[t!]
  \begin{center}
    \caption{Performace of XGBoost and Logistc Regression}
    \label{performance_1}
    \begin{tabular}{ccccc}
      \toprule
      Algorithm & Num of parties & Num of trees/epochs & Total time(s) & Average time(s) per tree/epoch\\
      \midrule
      XGBoost & 2 & 103 & 3743 & 36.34\\
      XGBoost & 3 & 111 & 2913 & 26.24 \\
      Logistic Regression & 2 & 2 & 1393 & 696.5 \\
      Logistic Regression & 3 & 2 & 1399 & 699.5 \\
      \bottomrule
    \end{tabular}
  \end{center}
 \end{table*}

 \begin{table*}[t!]
  \begin{center}
    \caption{Experimental results under different aggregation algorithms.}
    \label{performance_2}
    \begin{tabular}{ccccc}
      \toprule
      Model & Algorithm & Data Distribution  & Top-1 Acc & Num of Global Epochs(GEs) to reach high Acc \\
      \midrule
      \multirow{9}*{ResNet50} & FedAvg & \multirow{3}*{random} & 0.93 & \textbf{17 GEs to reach 0.90} \\
      ~ & FedProx & ~ & 0.93 & 23 GEs to reach 0.90 \\
      ~ & Scaffold & ~ &  \textbf{0.93} &  \textbf{17 GEs to reach 0.90} \\
      \cmidrule{2-5}
      ~ & FedAvg & \multirow{3}*{Dirichlet($\beta=0.5$)} & \textbf{0.91} & 18 GEs to reach 0.85 \\
      ~ & FedProx & ~ & 0.91 & 19 GEs to reach 0.85 \\
      ~ & Scaffold & ~ & 0.91 & \textbf{15 GEs to reach 0.85} \\
      \cmidrule{2-5}
      ~ & FedAvg & \multirow{3}*{Dirichlet($\beta=0.1$)} & 0.88 & 33 GEs to reach 0.80 \\
      ~ & FedProx & ~ & 0.88 & 24 GEs to reach 0.80 \\
      ~ & Scaffold & ~ & \textbf{0.88} & \textbf{13 GEs to reach 0.80} \\
      \midrule
      \multirow{9}*{VGG11} & FedAvg & \multirow{3}*{random} & 0.92 & \textbf{20 GEs to reach 0.90} \\
      ~ & FedProx & ~ & 0.91 & 24 GEs to reach 0.90 \\
      ~ & Scaffold & ~ &  \textbf{0.92} &  \textbf{20 GEs to reach 0.90} \\
      \cmidrule{2-5}
      ~ & FedAvg & \multirow{3}*{Dirichlet($\beta=0.5$)} & \textbf{0.90} & 17 GEs to reach 0.85 \\
      ~ & FedProx & ~ & 0.90 & 17 GEs to reach 0.85 \\
      ~ & Scaffold & ~ & 0.90 & \textbf{12 GEs to reach 0.85} \\
      \cmidrule{2-5}
      ~ & FedAvg & \multirow{3}*{Dirichlet($\beta=0.1$)} & \textbf{0.88} & 24 GEs to reach 0.80 \\
      ~ & FedProx & ~ & 0.87 & 20 GEs to reach 0.80 \\
      ~ & Scaffold & ~ & 0.87 & \textbf{11 GEs to reach 0.80} \\
      \bottomrule
    \end{tabular}
  \end{center}
 \end{table*}

\subsection{Privacy and Security}
 Privacy protection is the major concern of federated learning. But in most cases, it is neither practicable nor necessary to 
 protect the whole training process, which means a participant knows nothing except the output. Too strict privacy protection 
 usually results in low computing efficiency. XFL balances the security and efficiency in the distributed learning process, 
 by strictly protecting the input data and encrypting the intermediate data to ensure the sensitive privacy of every 
 participant cannot be accessed or infered by other participants. Besides, individual gradients are usually not allowed to be 
 exposed to prevent other parties or attackers from extrapolating the original data, no matter the rows or ids. 

 The federated algorithms in XFL is designed to meet the privacy requirements above. In terms of technology for privacy 
 protection, secure aggregation for horizontal federation consists of secure key exchange\cite{RFC7919}, secure presudo random 
 number generator\cite{SP800-90a} and one-time pad. Homomorphic encryption is supported for some vertical algorithms, MPC is also 
 in plan to support specific algorithms. All algorithms in XFL are based on the assumption that all parties are semi-honest, which 
 means they will excute the training process exactly but are curious to deduce some extra imformation. Collusion between parties 
 is not allowed for most of the algorithms.

\subsection{High Performace}
 Encryption algorithms are usually the performance bottleneck of federated learning. XFL adopts a number of self-developed methods 
 to improve the performance of cryptographic algorithms. For example, encryption by a security key in Paillier is much faster than 
 encryption by a public key. A DJN\cite{DJN} method for key generation and encryption has been utilized to further accelerate the 
 speed of the encryption. Fully homorphic encryption algorithm CKKS is applied to substitute Paillier for better performace, 
 because CKKS is more suitable for batch encryption. Moreover, a packing method is used for vertical XGBoost by packing two 
 plaintext to one to reduce nearly half of the calculation related to homomorphic encryption.

 On the other hand, we have also optimized the performance of computing and communication from the perspective of engineering 
 design. In Section \ref{Performance}, there are several experiments showing the performance of XFL.

\subsection{Supported Operators}
 We release many pre-built horizontal models such as VGG\cite{VGG}, ResNet\cite{ResNet} and BERT\cite{BERT}. NbAFL is an 
 experimental horizontal algorithm deploying differential privacy technology. XGBoost, logistic regression, K-means and other 
 algorithms are supported in vertical scenario. XFL supports not only conventional horizontal and vertical federated learning 
 algorithms, but also provides some local algorithms to help building a federation task. Non-federated algorithms such as 
 normalization, standard scalar are provided for local usage. The supported algorithms are listed in TABLE \ref{algorithms}. 
 More algorithms will be released in the future.

\section{Performance}\label{Performance}
 We demonstrate the efficiency of XFL by a series of experiments. In this part, we test the performance of vertical logistic 
 regression and XGBoost operators. The experimental dataset is Epsilon\cite{Epsilon}, where the training dataset contains 400000 
 samples and the validation dataset contains 100000 samples. The dataset consists of 2000 columns of features and 1 column of 
 label. Randomly, we choose 500 columns from the features and equally split them into 2 or 3 stacks, corresponding to 2 or 3 
 parties respectively. For each party, the basic hardware contains a 16-cores 2.3GHz CPU and a 256GB memory. The network 
 bandwidth is 1000Mb/s. We set a goal KS value on validation dataset for each operators, and record the total running time. The 
 goal KS value is 0.586 for XGBoost experiments, and 0.653 for logistic regression experiments. The main hyper-parameters in 
 XGBoost experiments are described as below: the tree depth equals to 2; the top rate of goss is 0.01; the other rate of goss is 
 0.02; the encryption method is Paillier and the key bit-length is 2048. In logistic regression experiments, the batch size is 
 2048 and the encryption method is CKKS. The results have been exhibited in TABLE \ref{performance_1}. The performance of both 
 the algorithms is outstanding in all existing frameworks.

 In vertical federations, there is usually no loss of accuracy. However, this is not the case in horizontal federations. A number 
 of studies have pointed out that the accuracy of joint modeling is lower than that of local modeling when the data is non-IID 
 (non-independent-and-identically distributed)\cite{FedAvg, Kairous2019, QYang2019}. Some experimentss\cite{QLi2022} have shown 
 different algorithms can improve the accuracy of horizontal federation with non-IID data, and the algorithms with excellent 
 performance has been implemented on XFL, such as FedProx and Scaffold. In this part, we conduct some experiments to show that 
 these algorithms can significantly improve the training efficiency and reduce the number of required global epochs.
 
 The typical horizontal federation learning in XFL mainly includes three steps: First, the server broadcast the global model to 
 all participants. Second, each participants conducts model training on the basis of local data. Finally, The server collects the 
 local models and aggregates them into a global model. Currently, XFL provides three aggregation methods: FedAvg, FedProx and 
 Scaffold. FedAvg is widely used because of its simplicity and efficiency, however, when the data in different participants is 
 non-IID, FedProx and Scaffold may achieve better results. To compare the above three aggregation algorithms when the data is 
 non-IID, we conduct experiments by utilizing the XFL framework. The experimental dataset is CIFAR-10\cite{CIFAR-10}, and the 
 models are ResNet50 and VGG11. There are two participants in each experiment. Each experiment contains 60 global epochs, and each
 global epoch contains 5 local epochs. The hyper-parameter of FedProx is fixed to $\mu = 0.005$. To achieve non-IID data, the 
 experiments adopt Dirichlet distribution method to split data. The smaller of $\beta$ means less coincidence of labels on each 
 side. The results are shown in TABLE \ref{performance_2}, where the bold font denotes the best performance. FedAvg always 
 performs well when the data is random distributed, i.e. the data is IID. However, its convergence rate becomes very slow when 
 the data is non-IID. In XFL, FedProx and Scaffold can be easily configured to achieve faster convergence. Especially, in this 
 group of experiments, Scaffold costs less than half of the global epochs of FedAvg to achieve the accuracy of 0.80.

\section{Conclusion}\label{Conclusion}
 With the ever increasing legislation on privacy protection and the ever larger need for data access, the demand for 
 privacy-preserving computing is growing stronger. By combining decentralized machine learning with cryptography, federated 
 learning has become one of the hottest privacy computing technologies. Nevertheless, the wide adoption of federated learning is
 stumped by technical and practical difficulties. XFL tries to solve this problem by providing a full-fledged, industry-grade, 
 easy-to-use and easy-to-develop federated learning library. XFL implements most of the popular federated learning 
 algorithms for both horizontal and vertical scenarios. It achieves high performance by using of mature architecture ready for 
 production. Also, to ease the development of new algorithms, the encryption, communication, and aggregation components of 
 federated learning has been made as plug-in-able as possible. With these design choices, XFL can be used both for experienced 
 machine learning engineers and novices who just want make their first federated learning application.

\end{document}